\documentclass[11pt]{article}
\usepackage{fullpage}

\usepackage[utf8]{inputenc} 
\usepackage[T1]{fontenc}    
\usepackage{hyperref}       
\usepackage{url}            
\usepackage{booktabs}       
\usepackage{amsfonts}       
\usepackage{nicefrac}       
\usepackage{microtype}      
\usepackage{epsf, setspace, latexsym, cite}
\usepackage{xspace, amsmath, amssymb, hyperref, bm}
\usepackage{color,amsthm}
\usepackage{graphicx}
\usepackage{color}
\usepackage{url}
\usepackage{subcaption}
\usepackage{graphicx}
\usepackage{algorithm}

\def\QED{\mbox{\rule[0pt]{1.5ex}{1.5ex}}}
\def\endproof{\hspace*{\fill}\tilde \QED\par\endtrivlist\unskip}

\newcommand{\1}[1]{\ensuremath{{\bf 1}_{\{#1\}}}}

\def\D{\ensuremath{{\mathcal D}}\xspace}

\def\R{\ensuremath{{\mathbb R}}\xspace}

\def\X{\ensuremath{{\mathcal X}}\xspace}
\def\Y{\ensuremath{{\mathcal Y}}\xspace}

\def\argmin{\ensuremath{\mbox{argmin}}}

\def\iid{\stackrel{i.i.d.}{\sim}}

\newcommand{\TD}{\mathrm{TD}}
\newcommand{\RTD}{\mathrm{RTD}}
\newcommand{\PBTD}{\mathrm{PBTD}}
\newcommand{\VCD}{\mathrm{VCD}}
\newcommand{\VS}{\mathrm{VS}}

\title{An Overview of Machine Teaching}
\author{
Xiaojin Zhu\\
\small UW-Madison\\
\small Madison, WI, USA\\
\small jerryzhu@cs.wisc.edu
\and
Adish Singla\\
\small MPI-SWS\\
\small Saarbr\"ucken, Germany\\ 
\small adishs@mpi-sws.org
\and
Sandra Zilles\\
\small Univ. of Regina\\
\small Canada\\
\small zilles@uregina.ca
\and
Anna N. Rafferty\\
\small Carleton College\\
\small Northfield, MN, USA\\
\small arafferty@carleton.edu
}
\date{\today}

\begin{document}
\maketitle

\begin{abstract}
In this paper we try to organize machine teaching as a coherent set of ideas.
Each idea is presented as varying along a dimension.
The collection of dimensions then form the problem space of machine teaching, such that existing teaching problems can be characterized in this space.
We hope this organization allows us to gain deeper understanding of individual teaching problems, discover connections among them, and identify gaps in the field.
\end{abstract}

\section{Introduction}

We start with several examples of machine teaching, with the goal of contrasting machine teaching with machine learning, in particular passive learning and active learning in supervised learning.
Consider learning a 1D threshold classifier where the input distribution $P_X$ is uniform over the interval $[0,1]$, the true threshold is $\theta^*$, and the binary label is noiseless: $y:=\theta^*(x)=\1{x\ge\theta^*}$:

\centerline{\includegraphics[width=0.7\textwidth]{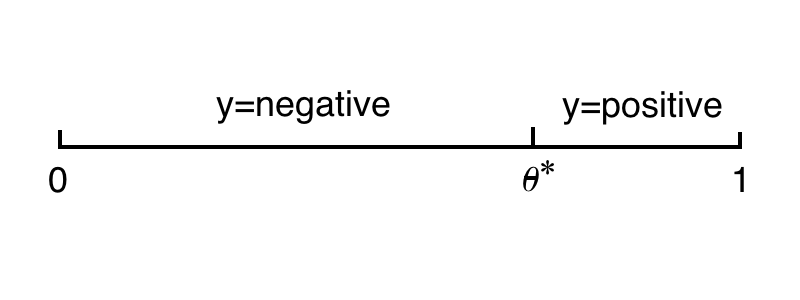}}

Passive learning receives $n$ training items 
$x_1, \ldots, x_n \iid P_X$ with 
$y_i = \theta^*(x_i)$.
It can be shown that with large probability a consistent learner (one that makes zero training error) incurs a generalization error 
$|\hat\theta - \theta^*| = O(n^{-1})$.
Intuitively with $n$ uniform training items the average spacing is $1/n$ which is the uncertainty of the decision boundary, as defined by the inner-most pair of negative, positive training items:

\centerline{\includegraphics[width=0.7\textwidth]{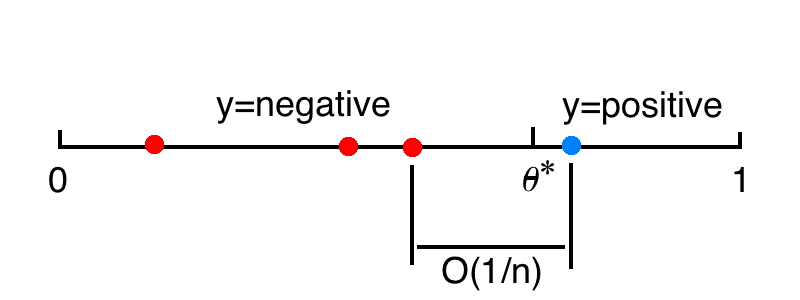}}

Equivalently, to achieve an $\epsilon$ generalization error one would need
$n \ge O(\epsilon^{-1})$ training items.
For example, if the desired generalization error is 0.001 the training set needs to be on the order of 1000.

Active learning allows the learner to adaptively pick queries $x$, and an oracle answers the label $\theta^*(x)$.
For our problem, active learning is equivalent to binary search on the interval $[0,1]$.
With each query, the learner can remove half of the remaining interval since it can deduce that the threshold cannot be in that half:

\centerline{\includegraphics[width=0.7\textwidth]{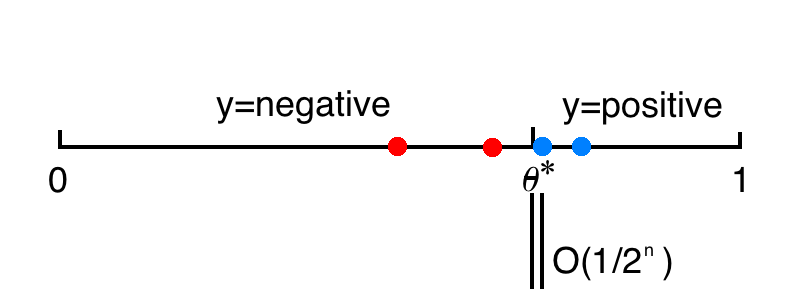}}

This results in a geometric reduction of generalization error
$|\hat\theta - \theta^*| = O(2^{-n})$.
Equivalently, to achieve an $\epsilon$ error the number of active learning queries is
$n \ge O(\log(\epsilon^{-1}))$.
For example, for the same 0.001 error an active learner only needs around 10 queries.

Machine teaching involves a teacher who knows $\theta^*$ and designs an optimal training set (also called a teaching set) for the learner (also called the student).
For any consistent learner, it is easy to see that the teacher can construct such a teaching set by judiciously picking two training items, one negative and the other positive, such that they are at most $\epsilon$ apart and contain $\theta^*$ in the middle.
The learner trained on this teaching set by definition achieves $\epsilon$ generalization error.
Importantly, the teaching set size is always two regardless of the magnitude of $\epsilon$:

\centerline{\includegraphics[width=0.7\textwidth]{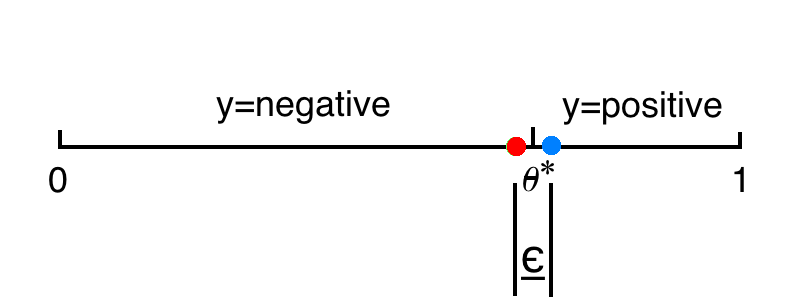}}

Let us look at a second example, where the teacher teaches a hard-margin SVM a target hyperplane decision boundary in $d$-dimensional space.
It turns out the teacher can again construct a teaching set consisting of two $d$-dimensional points.
In fact there are infinitely many such teaching sets, as long as the line segment between the two points is bisected by, and perpendicular to, the target hyperplane:

\centerline{ \includegraphics[width=2in]{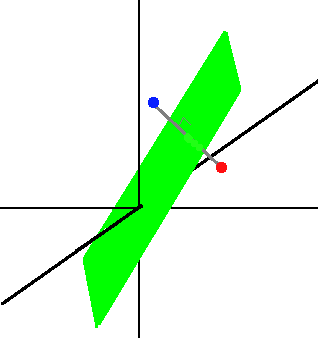} }

The so-called Teaching Dimension -- the smallest teaching set size --  of this problem is $\TD=2$.
This is to be contrasted with the VC dimension of $d$-dimensional hyperplanes, which is $\VCD=d+1$.

As a third example, consider a teacher who wants to teach a $d$-dimensional Gaussian density $N(\mu^*, \Sigma^*)$ to a learner.
The learner learns by computing the sample mean and sample covariance matrix of given data:
\begin{eqnarray}
\hat\mu &=& \frac{1}{n}\sum_{i=1}^n x_i \nonumber \\
\hat\Sigma &=& \frac{1}{n-1} \sum_{i=1}^n (x_i-\hat\mu) (x_i-\hat\mu)^\top. \nonumber
\end{eqnarray}
One can show that the teacher can construct a teaching set with 
$d+1$ points, which are the vertices of a $d$-dimensional tetrahedron centered at $\mu^*$ and scaled appropriately:

\centerline{\includegraphics[width=2in]{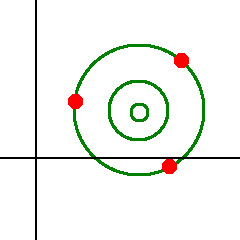}}

Let us contrast machine learning vs. machine teaching more formally, but still use passive learning as the problem setting.
Machine learning takes a given training set $D$ and learns a model $\hat\theta$.
The learner can take a variety of forms, including
version space learners for the theoretical study of Teaching Dimension (to be discussed later)~\cite{Goldman1995Complexity},
Bayesian learners~\cite{Zhu2013Machine},
deep neural networks,
or cognitive models which model how humans learn~\cite{Patil2014Optimal}.
For now, we will use the popular regularized empirical risk minimization framework as an example:
\begin{eqnarray}
\hat\theta = \argmin_{\theta \in \Theta} \sum_{(x,y)\in D} \ell(x,y,\theta) + \lambda \|\theta\|^2 
\label{eq:ERM}
\end{eqnarray}
where $\Theta$ is the hypothesis space, $\ell$ is the loss function, $D$ is the training set, and $\lambda$ is the regularization weight.

In contrast, in machine teaching the target model $\theta^*$ is \emph{given}, and the teacher finds a teaching set $D$ -- not necessarily $i.i.d.$ -- such that a machine learner trained on $D$ will approximately learn $\theta^*$. 
One special instance of machine teaching can be written as a bilevel optimization problem:
\begin{eqnarray}
\min_{D, \hat\theta} && \|\hat\theta - \theta^* \|^2 + \eta \|D\|_0  \\
\mbox{s.t.} && \hat\theta = \argmin_{\theta \in \Theta} \sum_{(x,y)\in {D}} \ell(x,y,\theta) + \lambda \|\theta\|^2.
\end{eqnarray}
Remarks:
\begin{itemize}
\item The upper optimization is the teacher's problem.  The teacher aims to bring the student model $\hat\theta$ close to the target $\theta^*$ while also to use a small teaching set ($\|D\|_0$ is the cardinality of the teaching set).
\item The teacher is typically optimizing over a discrete space of teaching sets.
\item The lower optimization is the learner's machine learning problem.
\item The teacher needs to know the learning algorithm to formulate this optimization.
\end{itemize}

\section{Why bother if the teacher already knows $\theta^*$?}
At this point we must address an important question: if the target model $\theta^*$ is known, what is the point of machine teaching?
One needs to go beyond the machine learning mentality.
There are applications where the teacher needs to convey the target model $\theta^*$ to a learner via training data.
For example:
\begin{itemize}
\item In certain education problems (with human students) the teacher has an educational goal that can be formulated as $\theta^*$.
For example, a geologist may want to teach students to categorize rocks into igneous, sedimentary, and metamorphic.
The geologist has the correct decision boundary $\theta^*$ in her mind, but she cannot telepathize $\theta^*$ into the student's mind.
Instead, she teaches by picking informative rock samples to show the students.
If the geologist has a good \emph{cognitive model} on how the students learn from samples, she can use machine teaching to optimize the choice of rock samples.
\item In one type of adversarial attack known as training-set poisoning, an attacker manipulates the behavior of a machine learning system by maliciously modifying the training data~\cite{Mei2015Machine,Alfeld2016Data,Mei2015Security}.
For example, 
consider a spam filter which constantly adapts its threshold in order to accommodate the changing legitimate content over time.
An attacker knowing the algorithm  may send specially designed emails to the spam filter to manipulate the threshold, such that certain spam emails can get pass the filter.
Here the attacker plays the role of the teacher, and the victim is the unsuspecting student.
\end{itemize}

In fact, it may be useful to understand machine teaching from a coding perspective~\cite{Zhu2015Machine}.
The teacher has a message which is the target model $\theta^*$. 
The decoder is a fixed machine learning algorithm $A$ which accepts a teaching set $D$ and decodes it into a model $A(D) := \hat\theta$.
The teacher must encode $\theta^*$ using the code words consisting of teaching sets. 
This is depicted in the following figure where each dot on the left is a teaching set.
Here the feasible code words are the preimage of $\theta^*$ under $A$, namely the black set of teaching sets.
The teacher would select among the preimage the smallest teaching set.

\centerline{\includegraphics[width=0.6\textwidth]{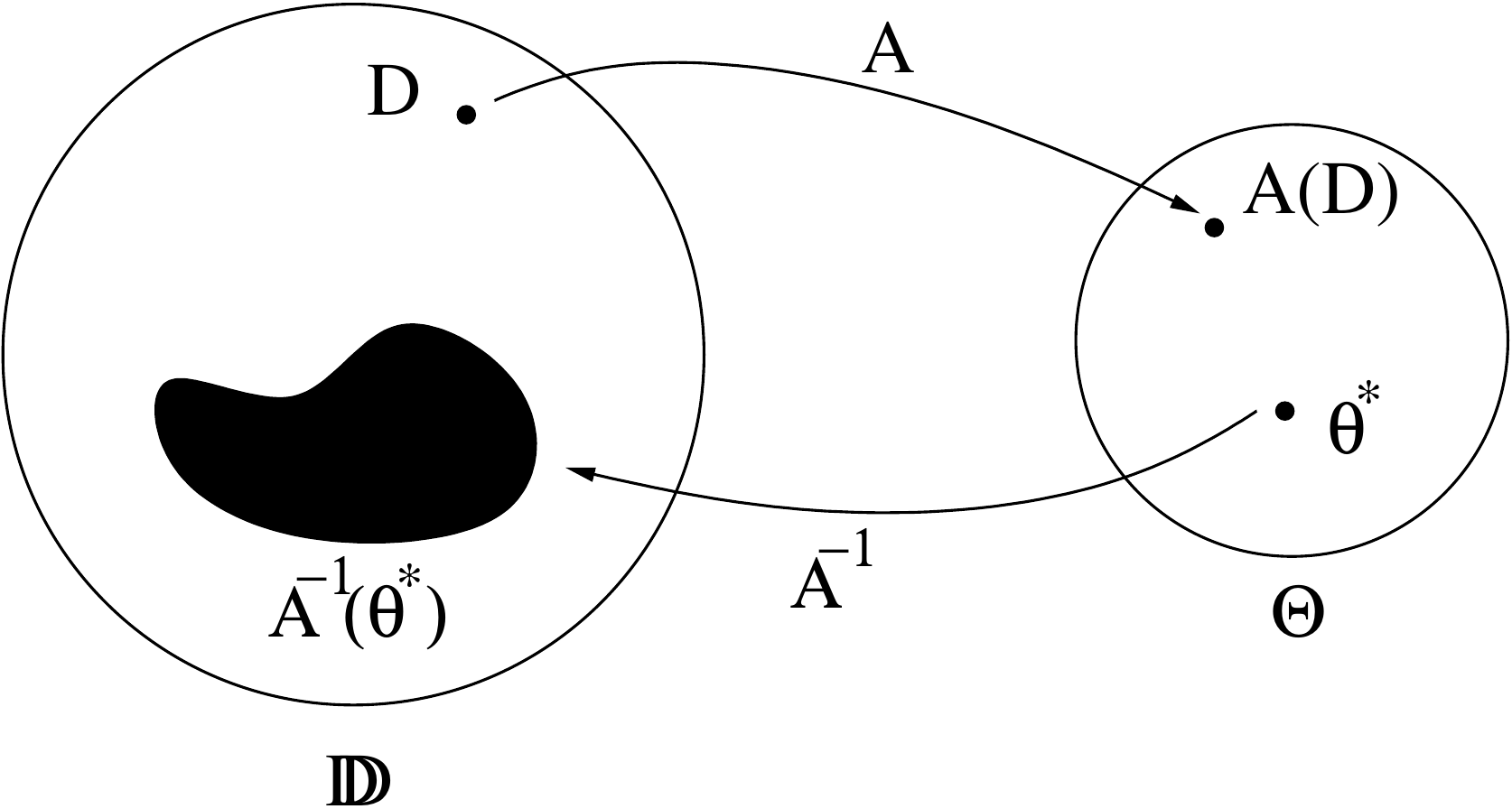}}

To illustrate the central role of the decoder which is the learning algorithm $A$, consider the task of teaching a homogeneous (no offset) linear regression function $y=\theta^* x$ for $x\in \R$.  One may think that a teaching set consisting of a single labeled example ``on the line'' $D_1=\{(x_1, \theta^* x_1)\}, x_1\neq 0$ is sufficient to teach.  This is indeed true, but only if the learner $A$ is the ordinary least squares (OLS) estimator: 
$A(D)=\argmin_\theta \|X\theta-y\|^2 = (X^\top X)^{-1}X^\top y$.
If, instead, the learner $A$ is ridge regression with regularization weight $\lambda$: 
$A(D)=\argmin_\theta \|X\theta-y\|^2 + \lambda\|\theta\|^2 = (X^\top X + \lambda I)^{-1}X^\top y$,
$D_1$ will not teach $\theta^*$ due to the learner shrinking the estimation toward zero.
The teacher can still teach with a single labeled example, but must nudge the $y$ value upward off the line in anticipation of learner shrinkage~\cite{JMLR:v17:15-630}:
$D_2=\{(x_1, \theta^* x_1 + \frac{\lambda \theta^*}{x_1})\}, x_1\neq 0$.
Note the amount of nudging depends on the property of the learner (decoder), specifically $\lambda$.
This also assumes that the teacher knows the decoder -- ways to relax this assumption are discussed in sections~\ref{sec:modelbasedfree} and~\ref{sec:multiplestudents}.

Therefore, machine teaching is not about learning a model (although that is used as a subroutine), but rather about generating data in order to transmit a model, control a learner, shape reinforcement learning, persuade an agent, influence vertices, or even attack an adaptive system.
It may be said that whenever one is optimizing data it is machine teaching; while if one is optimizing a model it is machine learning.
This leads us to a slightly more general formal definition of machine teaching:
\begin{eqnarray}
\min_{D, \hat\theta} && \mathrm{TeachingRisk}(\hat\theta)  + \eta \mathrm{TeachingCost}(D)  \label{eq:generalMT} \\
\mbox{s.t.} && \hat\theta = \mathrm{MachineLearning}(D). 
\end{eqnarray}
In the above, we define a generic function $\mathrm{TeachingRisk}(\hat\theta)$ for how unsatisfied the teacher is.  The target model $\theta^*$ is folded into the teaching risk function.  Alternatively, the teaching risk can be defined, e.g., by $\hat\theta$'s generalization error on an evaluation set and no target parameter $\theta^*$ is needed.

We also generalized the teaching cost function beyond the number of teaching items.
For example, different items may have different cognitive burden for a human student to absorb.

We can alternatively consider two constrained forms of the machine teaching problem.
One constrained form attempts to minimize the teaching cost subject to sufficient learning:
\begin{eqnarray}
\min_{D, \hat\theta} && \mathrm{TeachingCost}(D) \label{eq:mincoststrisk} \\
\mbox{s.t.} && \mathrm{TeachingRisk}(\hat\theta) \le \epsilon \\
&& \hat\theta = \mathrm{MachineLearning}(D). 
\end{eqnarray}
This form allows either approximate teaching or exact teaching (student must learn exactly $\theta^*$).
In fact, it includes classic Teaching Dimension as a special case of exact teaching.
Specifically, let $\mathrm{TeachingCost}(D)=\|D\|_0$ the cardinality of the teaching set, 
let $\mathrm{MachineLearning}(D)$ be the version space of the learner after seeing $D$,
and let $\mathrm{TeachingRisk}(\hat\theta)=0$ if $\hat\theta$ is the singleton set $\{\theta^*\}$ made from the target concept, $\infty$ for any other version space. 
The optimization objective is then precisely the Teaching Dimension of $\theta^*$ with respect to the hypothesis space $\Theta$, namely the minimum teaching set size to uniquely specify $\theta^*$.
More discussions are in section~\ref{sec:theory}.

The other constrained form allows a teaching budget and optimizes learning:
\begin{eqnarray}
\min_{D, \hat\theta} && \mathrm{TeachingRisk}(\hat\theta)  \\
\mbox{s.t.} && \mathrm{TeachingCost}(D)  \le B \\
&& \hat\theta = \mathrm{MachineLearning}(D). 
\end{eqnarray}

It should be noted that research on machine teaching is also relevant to scenarios in which the teacher has full information about the target, but does not have a succinct representation for it, i.e., the teacher knows everything about the behavior of the model but cannot formulate this knowledge in terms of model parameters $\theta^*$. In such cases, the teacher might still be able to provide the learning algorithm with carefully selected training examples, while the goal of the learning process would be to infer the model parameters. For instance, consider a scenario in which a human needs to extract certain information from web documents and would like to automate the extraction process. While the human may know exactly what type of information they are looking for, and can specify various highly representative examples of the desired information extraction, they may be incapable of formulating the machine-interpretable model that would be required for automating the information extraction process.

\section{Characterizing the machine teaching space}

We now organize the machine teaching space by introducing different dimensions.
Each teaching problem can then be thought of as a point or a region in this space.
This organization is by no means complete or prescriptive, and should be updated as the field moves along.

\subsection{The human vs. machine dimension: Who teaches whom?}
Machine teaching has the form ``teacher T teaches student S.''
While the mathematics behind machine teaching is unified, the applications can look quite different depending on who T, S are:
\begin{itemize}
\item T=machine, S=machine: A good example is data poisoning attacks.
S is a computer victim who runs a standard machine learning algorithm $A$ that learns model $A(D)$ from training data $D$.
T is an attacker -- a malicious program -- who can change (poison) $D$ and wants S to learn some nefarious $\theta^*$ instead. 
T also wants the poisoning to be subtle to avoid detection.
The attacking problem is a special case of machine teaching:
\begin{eqnarray}
\min_{\delta, \hat\theta} && \|\hat\theta-\theta^*\| + \eta \|\delta\| \\
\mbox{s.t.} && \hat\theta = A(D+\delta) 
\end{eqnarray}
with appropriate norms.
More discussions in section~\ref{sec:trustworthyAI}.

\item T=machine, S=human: An example is computer tutoring systems.
The key is to assume a cognitive model $A$ of the human student.
In the geology example earlier, $A$ may be the Generalized Context Model for human categorization.
T's goal is for the student to have high test scores in categorizing rocks, and T must teach by choosing training rock samples.
The computer tutor can pose this problem as~\eqref{eq:generalMT}.
More discussions in section~\ref{sec:education}.

\item T=human, S=machine: One good example is to utilize human domain experts to quickly train a text classifier. 
As the earlier 1D threshold classifier example demonstrates, a human teacher able to produce an optimal teaching set -- by either selecting documents from a corpus or even writing some new ones -- will vastly outperform active learning where the human is merely used by the machine as a label oracle.

Of course, the human teacher may not be optimal.
In that case, machine teaching allows the machine student to ``teach the human how to teach''~\cite{Suh2016Label}.
Take the same 1D example.
While the machine does not know the human teacher's target $\theta^*$, it knows the \emph{structure} of the optimal teaching set. 
Therefore, it can teach the human with analogues: ``If you want to teach $\theta^*=0.3$, you can show me a negative example at $x=0.29$ and a positive example at $x=0.31$.''

More broadly, human teachers afford the opportunity to teach beyond labeled examples: they can teach by features, pairwise comparisons, rules, etc. provided that the learning algorithm is equipped to accept such teaching signals.  
This is an active research direction in machine teaching.

\item T=human, S=human:
While traditional education is not the focus of machine teaching, insights gained from machine teaching have the potential to enhance pedagogy.
\end{itemize}

\subsection{The teaching signal dimension: What can the teacher use?}
The teaching signal depends on the learner.

For supervised learning, the standard setting is to teach with labeled training items.
We distinguish the following settings:
   \begin{itemize}
   
   \item synthetic / constructive teaching: 
   The teacher can use any item $x$ in the (typically continuous) feature space $\X$.
   For an $x\in \X$ the teacher typically needs to synthesize / construct an artificial item that has feature $x$, hence the name.
   We may further distinguish between an honest teacher, who is constrained to use the correct label according to the target model $\left(x \in \X, y=\theta^*(x)\right)$, and a lying teacher who is free to pair any label with $x$: $(x,y) \in \X \times \Y$.
   In education the latter may raise ethical questions.
   
   \item {pool-based teaching}:  
   For many applications it is desirable to teach with actual items, e.g. real images or documents.
   Artificial items may look nonsensical to a (human) student.
   It is therefore reasonable to assume that the teacher has a finite pool of candidate items, and must choose teaching items from the pool.
   This is the same consideration as pool-based active learning.
   
   \item 
   Of course, there can be hybrid settings.  For example, the teacher may start with a pool of candidate images.
   But for any image $x$ in the pool, the teacher can perform mild synthesis by scaling, rotating, and cropping the image.
   As another example, in training set poisoning the attacker often alters an existing training set $D_0$ by perturbing some training item $(x,y)$ into $(x+\delta_x, y+ \delta_y)$.
   
   \end{itemize}
Newer teaching signals for supervised learners are being studied.
Examples include teaching by features or pairwise comparisons.
Of course, this requires the ability of the learning algorithm to accept such signals.

For reinforcement learning, the teaching signal can be a demonstration in the form of a state trajectory. 
Or it can be artificial rewards in reward shaping.
Both aim to teach a policy to the reinforcement learning agent.

For maximizing memory recall in humans one teaching signal is ``spacing'': choosing when to prompt the human student to review an item.

\subsection{The batch vs. sequential dimension: Teaching with a set or a sequence?}
In batch teaching the teacher gives a training set to the student, which is a batch learner.  The order of training items does not matter.
Depending on the problem setting, the training set may or may not contain repeated items.

In sequential teaching the teacher must also optimize the order of training items, as the student is a sequential learner.
For example, some argue that it is important for the cognitive model of human students to be sequential, as humans are sensitive to the sequential effect.
In machine learning, sequential learners include Stochastic Gradient Descent (SGD) algorithms, multi-armed bandits (MAB), and reinforcement learners (RL). Teaching sequential learners is most commonly used in teaching robots, as in the ``teaching by demonstration'' approach~\cite{ArgallCVB09}. 
Curriculum learning~\cite{BengioLCW09} is a general approach in which training examples are presented to the learner in a sequence that is tailored to learning ``from simple to difficult.'' One can also show that perceptron-type learning has much better guarantees in terms of the convergence rate when examples far away from the decision boundary are presented before examples that are closer to the decision boundary~\cite{VembuZ16}.
Sequential teaching can also benefit the teacher if the teacher can obtain information about the learner’s state during the course of teaching (as discussed further in sections~\ref{sec:modelbasedfree} and \ref{sec:modelbasedfree}).
 
\subsection{The model-based vs. model-free dimension: How much does the teacher know about the student?}
\label{sec:modelbasedfree}
Machine teaching critically depends on the amount of knowledge the teacher has about the student.

On one extreme the teacher takes a model-based approach.  The student is a clearbox where the teacher has full knowledge of the learning algorithm.
For example,
\begin{itemize}
\item If the student is a Support Vector Machine (SVM), the teacher knows in~\eqref{eq:ERM} that $\ell()$ is the hinge loss and also the value of the regularization weight $\lambda$.
\item If the student is an SGD algorithm, the teacher knows the initial parameter $w_0$ of SGD, the loss function $\ell()$ or its gradient, and the learning rate parameter $\eta$.
\item In classic Teaching Dimension, the teacher knows that the student is a version space learner, who maintains a version space $\VS$ consisting of all hypotheses that are consistent with the training data $D$:
\begin{equation}
\VS(D) = \{\theta \in \Theta: \theta \mbox{ is consistent with } D\}.
\label{eq:VS}
\end{equation}
This also implies that the teacher knows the learner's hypothesis space $\Theta$ to start with. 
\end{itemize}
Such full knowledge of the student allows the teacher to specify teaching as the bilevel optimization problem~\eqref{eq:generalMT}.

On the other extreme the teacher takes a model-free approach.  
The teacher does not assume any learning algorithm used by the student.  Instead, the student is a blackbox to the teacher: the teacher gives it training data $D$ and only observes $\mathrm{TeachingRisk}$ as output.
This pointwise function evaluation view motivates the teacher to perform derivative free optimization on the student blackbox.

In between the two extremes is a graybox student, where the teacher assumes partial knowledge of the learning algorithm.
For example, the teaching may assume that the student is running ridge regression, i.e. the loss function $\ell$ in~\eqref{eq:ERM} is the squared loss; but the teacher does not know the value of regularization weight $\lambda$.
It is easy to see that, with uncertainty in $\lambda$, the teacher cannot exactly teach $\theta^*$ using any finite training set.
However, it may be possible for the teacher to ``probe'' the student.  For example, after teaching with a few items, the teacher may ask the student to predict on a new item $\tilde x$.  
The teacher may know that the student uses its current estimate $\hat\theta$ to make a noisy prediction:
$$\tilde y = \hat\theta^\top \tilde x + \epsilon$$
where $\epsilon$ is a zero-mean noise with known variance.  The teacher, upon receiving $\tilde y$ can then update its belief of the student's $\lambda$.
Such teaching / probing activities can be interleaved.
The teacher may benefit from applying active learning to such probing.

\subsection{The student awareness dimension: Does the learner know it is being taught?}\label{sec:student-awareness}
The vast majority of teaching settings involve a student that does not anticipate teaching.
The student may employ standard learning algorithms such as deep neural networks or reinforcement learning.
Typically (especially for supervised learning) the student assumes the training data is $i.i.d.$, oblivious of the fact that a teaching set may in fact be specially constructed and non-$i.i.d.$
The student simply applies the learning algorithm to that data and produces a model.
This is the assumption behind the machine teaching optimization problem~\eqref{eq:generalMT}.
In security applications, this means the victim does not anticipate attacks.

However, an increasing number of teaching settings now involve a student who is aware that it is being taught by a teacher.
\begin{itemize}
\item In computational learning theory, such anticipatory student enables various models of teaching, see, e.g.,~\cite{Balbach08,ZillesLHZ11,JMLR:v15:doliwa14a,gao2017preference}.  For example, the notions of Recursive Teaching Dimension ($\RTD$) and Preference-based Teaching Dimension ($\PBTD$) assume that the teacher and learner share a preference order over the class of all possible target concepts (i.e., the hypothesis space) and the learner expects the teacher to present examples that distinguish the target concept from all those that are ranked higher in that preference order. These models outperform the classical model of Teaching Dimension ($\TD$) in terms of the number of examples required for teaching~\cite{JMLR:v15:doliwa14a,gao2017preference}. Intuitively, the learner can identify the target faster if it knows the strategy according to which the given examples were chosen.

\item The student may understand that the teacher is a human who is suboptimal at teaching.  As discussed earlier, the student can educate the teacher on the structure of the optimal teaching set, thus improving the quality of the teaching set it receives.  The student may also decide to switch to active learning if it decides that the teacher is hopelessly suboptimal~\cite{cakmak2014eliciting,cakmak2014teaching,Suh2016Label}.

\item The student may be aware that it is running learning algorithm A, but the teacher is teaching for a different learning algorithm B.  
For example, A may be an SVM while the human teacher treats the computer learner more like a human child, where B is the implicit cognitive model of the child assumed by the teacher.
In this case, even if the teaching set is optimal for B it is in general suboptimal for A.
However, the student can ``translate'' the teaching set with the knowledge of A, B.
To illustrate, let $A$ be the ridge regression estimator with regularization weight $\lambda_A=2$:
$$A(D) = \argmin_{\theta \in \R} \sum_{(x,y)\in D} \frac{1}{2}(\theta x- y)^2 + {\lambda_A \over 2} \theta^2.$$
The teacher wants to teach the target $\theta^*=1$, but it assumes a slightly different learner B with $\lambda_B=1$.
The teacher could construct a singleton teaching set $(x=\theta^*,y=\lambda_B + x^2)=(1,2)$.
If A were to take this teaching set directly, it will learn a wrong $\hat\theta=\frac{xy}{x^2+\lambda_A}=\frac{2}{3}$.
Instead, if A is aware of the teacher's student model B, it could perform the following translation of the teaching set: 
$$\tilde x = \frac{xy}{x^2+\lambda_B}, \;\;\;
\tilde y = \lambda_A + {\tilde x}^2$$
The translated teaching set is $(1,3)$, making $A$ learn the correct $\theta=1$.

\item In security applications, this means the victim may employ defense mechanisms to resist attacks~\cite{Alfeld2017Explicit}.

\item In educational applications, a human student may expect that the teacher is selecting examples specifically to aid learning (as discussed further in section~\ref{sec:education} )
\end{itemize}

\subsection{The one vs. many dimension: how many students are simultaneously taught?}
\label{sec:multiplestudents}
Most teaching settings have one teacher and one student.

In some settings, however, there may be many students with different learning algorithms.
This is the case, for example, in classroom teaching.
The key constraint is that the teacher must use the \emph{same} teaching set on all students.
Even if the teacher has perfect knowledge of each student's learning algorithm, it is in general impossible to perfectly teach all students~\cite{Zhu2017NoLearner}.
The teacher may choose to optimize teaching for the worst student, which leads to a minimax risk formulation:
\begin{eqnarray}
\min_{D, \{\hat\theta_\lambda\}} && \max_{\lambda} \mathrm{TeachingRisk}(\hat\theta_\lambda)  + \eta \mathrm{TeachingCost}(D) \\
\mbox{s.t.} && \hat\theta_\lambda = \mathrm{MachineLearning}_\lambda(D)
\end{eqnarray}
where the subscript $\lambda$ denotes different students.
Alternatively, the teacher may choose to optimize teaching for the average student, which requires a prior distribution $f(\lambda)$ (e.g. uniform) over the students and leads to the Bayes risk formulation:
\begin{eqnarray}
\min_{D, \{\hat\theta_\lambda\}} && \int_{\lambda} f(\lambda) \mathrm{TeachingRisk}(\hat\theta_\lambda) d\lambda  + \eta \mathrm{TeachingCost}(D) \\
\mbox{s.t.} && \hat\theta_\lambda = \mathrm{MachineLearning}_\lambda(D).
\end{eqnarray}

\subsection{The angelic vs. adversarial dimension: Is the teacher a friend or foe?}
Machine teaching applications can be characterized by their intention.
This ranges from angelic (optimized, personalized education; 
improving cognitive models;
fast classifier training;
debugging machine learners~\cite{Zhang2018Training,Cadamuro2016Debugging,Ghosh2016Trusted}, etc.)
to adversarial (training-set poisoning attacks).

\subsection{The theoretical vs. empirical dimension: What is the work style?}
While most teaching problems have both theoretical and empirical components,
usually one of them is emphasized that dictates the approach toward the problems.
On one extreme, there is pure theoretical research centered on understanding the Teaching Dimension and its variants such as 
Recursive Teaching Dimension and Preference-based Teaching Dimension etc. 
On the other extreme, many computer tutoring systems employ heuristic teaching methods in order to improve human student performance.

\section{Some research directions in machine teaching}

\subsection{Algorithmic teaching theory}
\label{sec:theory}

A teaching setting where the most theoretical advances have to be made is the study of Teaching Dimension. One way to cast Teaching Dimension in the machine teaching framework is constrained optimization~\eqref{eq:mincoststrisk},
where the student is the version space learner~\eqref{eq:VS}, the teaching risk constrains the version space to be the singleton set $\{\theta^*\}$, and the teaching cost is the cardinality of the teaching set.

In some cases, this notion of teaching seems intuitively rather weak, but when designing different teaching models, one faces the issue of not trivializing the learner's role by introducing ``unfair coding tricks'' or ``collusion.'' For instance, for a countable hypothesis space over a countable domain, it is not desirable to teach the $i$th concept in the hypothesis space with the $i$th element of the domain (irrespective of its label), which would result in a teaching dimension of 1 for all such hypothesis spaces, assuming teacher and learner agree on specific enumerations of the hypothesis space and the domain. There is no generally adopted notion of collusion, see~\cite{GoldmanM96,ZillesLHZ11} for some examples. The Recursive Teaching Dimension ($\RTD$) and the Preference-Based Teaching Dimension ($\PBTD$) correspond to two notions of teaching that are collusion-free in the most stringent sense defined in the literature and that overcome many of the shortcomings of the classical $\TD$ approach. A systematic study of the effect of various notions of collusion is an interesting direction for future research.

One of the fundamental theoretical questions is whether there is any relationship between the information complexity of teaching (such as the $\TD$ or its variants) and the sample complexity of passive learning (particularly the VC Dimension ($\VCD$)) for a given concept class. While there is no general relationship between $\TD$ and $\VCD$, i.e., $\TD$ can be arbitrarily smaller and arbitrarily bigger than $\VCD$, recent results on the parameters $\PBTD$ and $\RTD$ suggest that teaching complexity is indeed related to the complexity of passive learning. While $\RTD$ can be arbitrarily smaller than $\VCD$, it was shown that $\RTD\in O(\VCD^2)$~\cite{HuWLW17}, and for a few special cases $\RTD$ is known to be equal to $\VCD$~\cite{doliwa2014recursive}. These results immediately transfer to $\PBTD$, since $\PBTD\le\RTD$ holds in general~\cite{gao2017preference}. An open question is whether $\RTD\in O(\VCD)$, i.e., whether $\RTD$ is upper-bounded by a function linear in the $\VCD$~\cite{SimonZ15}. One reason why this question is of interest to a wider audience within the computational learning theory community is that it appears to be related to the  \emph{sample compression conjecture}, which has been open for 30 years and which states that every concept class $\mathcal{C}$ has a ``sample compression scheme'' in which each sample set $S$ consistent with a concept in $\mathcal{C}$ can be compressed to a subset of size at most the $\VCD$ of $\mathcal{C}$ without losing any label information~\cite{LittlestoneW86,FloydW95}. Due to a number of results that show how to use teaching sets as compression sets and vice versa~\cite{DarnstadtKSZ16,doliwa2014recursive}, resolving the question whether or not $\RTD\in O(\VCD)$ would shed some light on the sample compression conjecture.

Teaching Dimension can be generalized to non-version space learners, such as ridge regression, logistic regression, and SVMs~\cite{JMLR:v17:15-630}, or Bayesian learners~\cite{Zhu2013Machine}.

The current notions of Teaching Dimension are based on a setting where teacher only provides labeled examples as input. It would be interesting to consider settings with different teaching signals. For instance, let us say a teacher can provide features of an example in addition to the label. Also, it would be interesting to consider new query modalities, for instance, teaching based on pairwise comparison queries instead of labels, cf. \cite{kane17active}.

Currently, most of the teaching models are studied in a batch setting where the teacher provides a set of examples in a batch to the learner. However, in real-world scenarios (e.g. personalized education), the teaching process is often interactive where the teacher can feed examples sequentially to the student, and can adapt these examples based on the current performance and feedback received from the student. One of the primary research questions is to understand how much speed up can be gained by adaptivity, cf. \cite{golovin2011adaptive,singla2013actively,singla14near,liu17iterative}. 

Another interesting algorithmic question inspired from real-world scenarios is to model students with limited memory and/or limited computational power---current models usually do not put any such constraint on the student. For instance, it would be important to understand the Teaching Dimension of different concept classes when teaching such limited-capacity students. It would also be interesting to consider more realistic scenarios of how to model this  limited capacity of students, cf. \cite{Patil2014Optimal}. 

\subsection{Human Robot/Computer Interaction}

In the context of improving the efficiency of human robot interaction, an exciting new direction of research is to study machine teaching formulation for reinforcement learning agents. This, in turn, can be useful in two ways: (i) insights gained here can be used to teach human users on how to optimally interact with and provide instructions to robots (e.g. for personalized robotic devices like Siri or Alexa)~\cite{DBLP:journals/corr/SimardACPGMRSVW17,akgun2012trajectories,thomaz2009learning}, (ii) we could develop smarter learning algorithms for robots that can anticipate being taught by a human teacher (see discussions in section~\ref{sec:student-awareness}).

One concrete setting is to study the problem of teaching a student that uses an inverse reinforcement learning (IRL) algorithm. In IRL, the student has access to demonstrations provided by an expert (\emph{e.g.}, trajectories when executing an optimal policy in an MDP),  and the goal is to learn a reward function using these demonstrations. In classic IRL, these trajectories are usually provided in a random fashion where the intention is not to teach (see the discussion of ``doing'' vs. ``showing'' in \cite{ho2016showing}). Here, one could study the optimization problem from the teacher's point of view on how to generate the best set of demonstrations for the student, cf. \cite{cakmak2012algorithmic} for initial results in this direction. Furthermore, there are several more challenging real-world scenarios, e.g. when there is a mismatch of the state-space in the underlying MDP (e.g. human and robot have a different view of the world given different sensory inputs), or when there is a mismatch in the perceived rewards. 

Another concrete setting is to study optimal reward shaping.  Reward shaping refers to a teacher manipulating the rewards delivered to a reinforcement learning agent, with the goal to quickly teach a target policy to the agent~\cite{ng1999policy}.
Optimal reward shaping aims to minimize the learning time and the total amount of reward manipulation. 

\subsection{Education}
\label{sec:education}
One exciting application domain for machine teaching is personalized education; here, machine teaching formulations could enable us further development of rigorous algorithms for intelligent tutoring systems. These settings frequently provide the machine teacher with information about the learner and/or the progress of learning via the learner's responses to a problem; this makes a sequential approach to machine teaching appropriate both because humans are often sensitive to ordering, and because the teacher gains more information over time.  While simple, potentially suboptimal, ways of deciding what data to present to the human learner are commonly used in automated teaching systems (e.g., ask the learner to solve a problem about any unmastered concept), there has been increasing recent interest in treating the problem as one of optimal control, often with approximate solutions that take into account learner responses~\cite{rafferty2016faster,whitehill2017approximately}. These approaches thus employ a teaching policy rather than seeking a fixed teaching set. One challenge in this area is because most approaches are model-based, accurate cognitive models must be identified for different learning tasks relevant to education. This has been an ongoing enterprise within cognitive and educational psychology, with increasing interest in creating or refining cognitive models based on existing datasets (e.g.,~\cite{lindsey2014automatic,Patil2014Optimal}). Behavioral experiments have also been employed within large scale MOOCs and intelligent tutoring systems; see \cite{Zhu2015Machine} and \cite{lindsey2013optimizing} for more discussion.

One concrete learning task that has gained a lot of recent interest is that of reviewing content via flashcards (e.g. for teaching vocabulary of a foreign language in Duolingo) \cite{settles2016trainable}. Given a set of vocabulary words that a student aims to learn, software tools (online websites or smartphone apps) for flashcards would shuffle through these cards with a goal to improve the recall probability of these words.  Here, the key is to properly model the forgetting aspect of learning, and many software tools are using spaced-repetition models based on a cognitive model that uses a parametric decay function on a recall probability variable. This is then an optimal control problem, where the control is when/how often the algorithm should perform a review/test action. There are several exciting questions here, for instance, how to learn parameters of the cognitive model on the fly (e.g. via exploration-exploitation techniques) in order to provide personalized reviewing schedule. One approach has been to leverage a probabilistic model to make inferences about the unknown difficulty of words and unknown individual differences among students based on distributions over the populations, enabling personalized review that improves as more students interact with the system~\cite{lindsey2014improving}. 
Another approach poses the spaced-repetition problem as stochastic differential equations and solves the optimal control problem~\cite{2017arXiv171201856T}.
Further exploration about the best ways to improve models while teaching and balance performance for an individual and for a population is needed.

While the most common approaches in education focus on a single learner and assume the learner is not sensitive to being taught, there have been several interesting explorations of other parts of the space delineated by the machine teaching dimensions above. For example, consideration of the best teaching set for a group of learners, each with somewhat varying parameters in their cognitive model, can lead to predictions about the best type of student groupings and sizes for instruction (e.g., are small class sizes or similar ability students best for helping all children learn?)~\cite{frank2014modeling,Zhu2017NoLearner}. Developmental psychology has also identified many cases where children are sensitive to whether the setting is pedagogical - that is, whether the informant is attempting to teach them. This has lead to automated teaching approaches that select examples for a learner who assumes a helpful teacher~\cite{shafto2014rational}. Examining how sensitive people are to helpful teachers across a range of domains is likely to lead to new approaches for a range of learning algorithms.

\subsection{Trustworthy AI}
\label{sec:trustworthyAI}

As discussed above, machine teaching can also be applied to an adversarial attack setting where the teacher is an attacker.
To elaborate, we distinguish two levels of adversarial attacks.
Level 1 adversarial attacks are test-time attacks.  They manipulate a test item $(\tilde x,\tilde y)$ such that a \emph{fixed, deployed} model $\hat\theta: X\mapsto Y$ would misclassify it.  Specifically, level 1 attacks find a small perceptual perturbation to change $\tilde x$ (e.g. a stop sign image) into $x$ (the stop sign image will a few pixels changed) so that $x$ is classified differently (e.g. into a yield sign) by $\hat\theta$.
This can be expressed as the following optimization problem:
\begin{eqnarray}
\min_{x} && \|\tilde x-x\|_p \\
\mbox{s.t.} && \hat\theta(x) \neq \tilde y, 
\end{eqnarray}
where the $p$-norm is a surrogate to perceptual distance.

In contrast, level 2 adversarial attacks are training-set poisoning attacks.
The inputs are
$D_0$ the original training set,
$A: \D \mapsto \Theta$ the learning algorithm,
and postcondition $\Psi: \Theta \mapsto Boolean$. 
Level~2 attacks find small perceptual perturbation to the training set $D_0$ so the trained model satisfies the postcondition:
\begin{eqnarray}
\min_{D} && \|D_0-D\|_p  \\
\mbox{s.t.} && \Psi(A(D)).
\end{eqnarray}
For instance, the postcondition can be $\Psi(\hat\theta) = [\hat\theta(\tilde x) = \bar y]$ which plants an attacker-desired classification $\bar y$ for test item $\tilde x$ implicitly through the poisoned training data and the training process.
Because the learning algorithm $A$ is involved, level 2 attacks are more challenging to solve than level~1 attacks.

Understanding the optimal training-set poisoning attacks can, in turn, help us design optimal defenses against attackers. For instance, we can develop an automated defense system that could flag the parts of training
data which are likely to be attacked (based on our model of the teacher) and focus human analysts' attention on those parts. This could drastically increase the chance of detecting such attacks by analysts once they know where to look, see \cite{Mei2015Machine,Zhang2018Training} for more discussions. 

An important line of research here would be to model interactions between an attacker (the teacher) and a learning algorithm (the student) as a repeated game. Here, we would like to design a robust learning algorithm that can anticipate about the teacher's actions (i.e. the future attacks) and develop an optimal forward-looking defenses against such attacks.
   
  
\subsection{Efficiently finding optimal teaching solutions}
The optimization problem for machine teaching~\eqref{eq:generalMT} is intrinsically hard due to its
combinatorial, bilevel nature.
Even simple instances are NP-hard: one can show that simple teaching problems include the set-cover and subset sum problems.
While some special cases have closed-form solution, 
many more require careful formulation.
For problems when the teaching set size is small, it is possible to formulate teaching as a mixed integer nonlinear programming (MINLP) problem and use existing solvers.
Some other problems can benefit from approximate algorithms with guarantees, for example, by utilizing the submodularity properties of the problems.

\bibliographystyle{plain} 
\bibliography{overview} 
\end{document}